\documentclass{article}
\usepackage{ijcai20}

\usepackage[normalem]{ulem}

\usepackage{times}

\usepackage{soul}
\usepackage{url}
\usepackage{hyperref}
\hypersetup{colorlinks,allcolors=black}
\usepackage[utf8]{inputenc}
\usepackage[small]{caption}
\usepackage{graphicx}
\usepackage{amsmath}
\usepackage{booktabs}
\title{Fine-grained Financial Opinion Mining: A Survey and Research Agenda}

\author{
Chung-Chi Chen$^{1}$
\and
Hen-Hsen Huang$^{2,3}$
\and
Hsin-Hsi Chen$^{1,3}$
\affiliations
$^1$Department of Computer Science and Information Engineering, National Taiwan University, Taiwan\\
$^2$Department of Computer Science, National Chengchi University, Taiwan\\
$^3$MOST Joint Research Center for AI Technology and All Vista Healthcare, Taiwan\\
\emails
cjchen@nlg.csie.ntu.edu.tw,
hhhuang@nccu.edu.tw,
hhchen@ntu.edu.tw
}

\begin{document}

\maketitle

\begin{abstract}
Opinion mining is a prevalent research issue in many domains.
In the financial domain, however, it is still in the early stages.
Most of the researches on this topic only focus on the coarse-grained market sentiment analysis, i.e., 2-way classification for bullish/bearish.
Thanks to the recent financial technology (FinTech) development, some interdisciplinary researchers start to involve in the in-depth analysis of investors' opinions.
In this position paper, we first define the financial opinions from both coarse-grained and fine-grained points of views, and then provide an overview on the issues already tackled.
In addition to listing research issues of the existing topics, we further propose a road map of fine-grained financial opinion mining for future researches, and point out several challenges yet to explore. 
Moreover, we provide possible directions to deal with the proposed research issues.
\end{abstract}

\section{Introduction}
\label{section introduction}
Dealing with the data in the financial domain is one of the hot research directions in the artificial intelligence (AI) community.
Following the recent trend of financial technology (FinTech), several workshops are held in conjunction with major conferences such as FinNLP~\cite{ws-2019-financial}, ECONLP~\cite{emnlp-2019-economics}, FNP~\cite{ws-2019-financial-narrative}, and DSMM~\cite{10.1145/3336499}.
This reflects the increasing interest of the AI researchers in financial and economic domains.
The special track in IJCAI-2020, AI in FinTech, also evidences this phenomenon.

Recently, more and more interdisciplinary cooperation between finance and computer science, and interesting research results are published.
Some works~\cite{sedinkina-etal-2019-automatic,qin-yang-2019-say,yang2020html} introduce the earning conference call, which is one of the important meetings for announcing the news of a company, to the natural language processing (NLP) community.
Some works~\cite{maia201818,chen2019crowd} pay their attention to financial social media data, and propose novel tasks for in-depth investigations.
These works indicate the trend of fine-grained opinion mining in the financial domain.

When mentioning the opinion in Finance, bullish/bearish comes into most people's minds.
However, the market sentiment of the financial instrument is just one kind of opinions in the financial industry.
As other industries such as manufacturing and textiles have many kinds of products, there are also a lot of products in the financial industry.
Financial service is also a major business of many financial companies, especially, in the recent FinTech trend. 
For instance, many commercial banks focus on the business of both loan and credit card.
Although many issues could be explored in the financial domain, most researchers in the AI and the NLP communities only focus on the market sentiment of the financial instrument. 
In this paper, we sort out several research issues that can broaden the research topics in the AI community. 

This paper is aimed at providing an overview of where we are in fine-grained financial opinion mining and helping the community to understand where we should be in the future.
For understanding the past and the present works, we discuss the components of the financial opinions one-by-one with related works.
During the discussion, we will point out some possible research issues.
For the future research directions, we mainly focus on illustrating the unexplored challenges.
We provide a research agenda with the directed graphs toward financial opinions.

At beginning, some concepts need to be declared.
Firstly, market profits are rather aleatory, so that they cannot be used as labels of opinions, and those studies focusing on constructing end-to-end models for market movement prediction~\cite{hu2018listening,feng2019enhancing} will not be included in this paper.
Secondly, news articles describing the events do not contain opinions.
Thus, those studies analyzing the events in news articles~\cite{zheng-etal-2019-doc2edag} will not be covered by this paper.
Thirdly, following the investigations of previous works~\cite{loughran2009liability,chen2018ntusd,chen2020issues}, general sentiment (positive/negative) are different from market sentiment (bullish/bearish).

This paper is organized as follows. Section~\ref{section related work} compares this paper with previous surveys. Section~\ref{section definition} provides careful definitions of coarse-grained and fine-grained financial opinions.
Section~\ref{section opinion mining} lists the existing tasks and points out critical research issues.
Section~\ref{section road map} proposes novel extension tasks of financial opinion mining.
Section~\ref{section Blueprint} provides potential directions for the extension tasks.
Finally, Section~\ref{section conclusion} concludes remarks.

\section{Related Work}
\label{section related work}
Pang and Lee~\shortcite{pang2008opinion} and Liu~\shortcite{liu2015sentiment} provide a general overview of sentiment analysis and opinion mining.
Then the overview and survey papers related to opinion mining in the general domain are updated every year~\cite{pradhan2016survey,abirami2017survey,hussein2018survey,tedmori2019sentiment}.
Most of the works focus on the opinions on social media platforms~\cite{kharde975sentiment,li2019survey,soong2019essential}.
Some works focus on specific topics such as product review~\cite{jebaseeli2012survey} and reputation evaluation~\cite{chiranjeevi2019survey}.
Although Kumar and Ravi~\shortcite{kumar2016survey} provide a survey on text mining in finance, few of the previous work provides an arrangement of opinion mining in finance.
In this paper, we will formulate the financial opinion mining task and illustrate a big picture of this research area.

Although some previous surveys have paid their attention to text mining in finance~\cite{das2014text,fisher2016natural}, they show less solicitude for opinion mining, and only mention a few coarse-grained financial opinion mining tasks.
In order to provide an in-depth look at the recent trend---fine-grained financial opinion mining, this paper mainly focuses on the issues of fine-grained financial opinion mining and proposes a road map for the future research.

\begin{figure}[t]
    \centering
    \includegraphics[width=8cm]{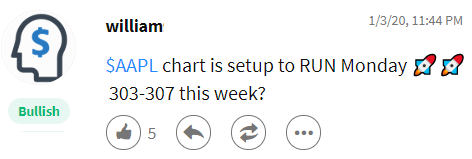}
    \caption{Example of investor's opinion.}
    \label{fig:Example of investor's opinion}
\end{figure}

\section{Financial Opinion Definition}
\label{section definition}
In this paper, we separate the financial opinions into two parts: (1) investor's opinion, i.e., the opinion about financial instruments and (2) customer's opinion, i.e., the opinion about the financial product or financial service.
As we mentioned in Section~\ref{section introduction}, market sentiment and general sentiment are different, and the previous works~\cite{loughran2009liability,chen2018ntusd,chen2020issues} already evidence this claim.
Therefore, when discussing the investor's opinion, the sentiment is the market sentiment.
On the other hand, when discussing the opinion of a financial product or a financial service, the sentiment denotes the general sentiment.
Based on these two opinion types, we define the opinions by both coarse-grained and fine-grained viewpoints.

\subsection{Coarse-grained Financial Opinion}
As the opinion mining task in other domains, the coarse-grained financial opinions can be separated into two (bullish/bearish or positive/negative) or three (bullish/bearish/neutral or positive/negative/neutral) classes, and each opinion related to one target entity.
In most cases, the opinion holder and the publishing time are given.
All of the above information can construct a 4-tuple to represent a coarse-grained opinion: 
\begin{quote}
    $(e,s,h,t^p)$\,,
\end{quote}
where $e$ denotes the target entity, $s$ denotes the sentiment, $h$ denotes the opinion holder, and $t^p$ denotes the publishing time.

Figure~\ref{fig:Example of investor's opinion} shows an example on one of the famous financial social media platforms, StockTwit\footnote{\url{https://stocktwits.com}}.
The 4-tuple of this tweet is 
\begin{quote}
    (\$AAPL, Bullish, william, 1/3/20 11:44PM)
\end{quote}
Because all essential terms are provided by the platform and users, researchers can easily collect lots of labeled data from the platform.
Therefore, many previous works adopt this dataset to construct a market sentiment lexicon for financial social media~\cite{oliveira2016stock,li2017learning,chen2018ntusd}, and lots of previous works use the labels to test their sentiment analysis models.
Renault~\shortcite{renault2019sentiment} provides a comparison between different models on the financial social media data.

Few works involve in mining the opinions of the products and the service from customers in the financial domain, although these are also very important for financial institutions.
For example, customer satisfaction and customer service quality are the focuses in the financial domain~\cite{potluri2016structural,ali2017service,tomar2019comparative}.
Capturing the opinions from customers' discussion on the online forum is a possible direction for evaluating customer satisfaction and customer service quality.
On the other hand, the opinions of the products can also provide cues for improving the next product.
For example, in the discussion of a credit card forum, the reply like (\textbf{E1}) shows the opinion on the credit card, called FlyGo.
\vspace{5pt}

\noindent \textbf{Example (E1)}: 

\textit{Because the cashback of FlyGo was canceled, I cut it directly.}

\vspace{5pt}

\noindent The 4-tuple of this reply is 
\begin{quote}
    (FlyGo, Negative, CREA, 12/31/19 21:04)
\end{quote}

Based on the above issues, we list the following research questions:

\vspace{5pt}
\noindent (\textbf{RQ1}) How to identify the opinions toward a specific product or a specific service in financial industry from the online forums or social media platforms?

\noindent (\textbf{RQ2}) To what extent the opinions on the online forums impact customer satisfaction or the sales of the products?    

\noindent (\textbf{RQ3}) What can or cannot be transferred from the existing opinion mining task in other domains to financial domain?

\subsection{Fine-grained Financial Opinion}
In this section, we add the related components one-by-one to define the fine-grained financial opinion. 
The first one is the aspect of the opinion.
Still, we start from the investor's opinion.
Taking the tweet in Figure~\ref{fig:Example of investor's opinion} as an example, the analysis aspect is technical analysis, because the analysis is based on the price chart.
Thus, we extend the 4-tuple to a 5-tuple as follows:
\begin{quote}
    $(e,s,h,t^p,a)$\,,
\end{quote}
where $a$ denotes the analysis aspect.
Maia et al.~\shortcite{maia201818} and our previous work~\shortcite{chen2019numeral} provide datasets for extracting the analysis aspect of the investors.

The other important component is the degree of sentiment.
Adding the degree of sentiment ($d$) to the 5-tuple, we get a 6-tuple as follows:
\begin{quote}
    $(e,s,h,t^p,a,d)$
\end{quote}
Some works in other domains extend the sentiment into five classes based on the strongness~\cite{balikas2017multitask,akhtar2019all}.
In the financial domain, Cortis et al.~\shortcite{cortis2017semeval} label the degree of sentiment into the range between $-$1 and 1.

A fine-grained customer's opinion can also be shown as a 6-tuple.
With this definition and the five-class degree setting, we can extend (\textbf{E1}) to:
\begin{quote}
    (FlyGo, Negative, CREA, 12/31/19 21:04, cashback, 5)
\end{quote}

Since few works discus the customer's opinion, the following research questions are still unexplored:
\vspace{5pt}

\noindent (\textbf{RQ4}) How to define the aspects for the financial products and financial services?

\noindent (\textbf{RQ5}) What kind of evaluation method is proper for the customer's opinions?

\vspace{5pt}

In the example in Figure~\ref{fig:Example of investor's opinion}, the investor claim that the price of \$AAPL will in the range of 303 to 307 in the coming trading days (one week).
In this case, a financial opinion can be extended to a 7-tuple:
\begin{quote}
    $(e,s,h,t^p,a,d,C)$\,,
\end{quote}
where $C$ denotes the set of investor's claims.
Note that a big difference exists between the investor's claim and the customer's claim.
Different from customers, who provide their opinions based on the experience of using a product or the service in the past, investors provide their opinions for future events based on their analysis.
In our previous work~\cite{chen2019crowd}, we analyze five kinds of investor's claims, which contain the price information.
We will detail it in Section~\ref{section opinion mining}.
Here raises a research question:
\vspace{5pt}

\noindent (\textbf{RQ6}) How to detect the claims related to the financial opinion?
\vspace{5pt}

The other characteristic of the investor's opinion is that the market information of the financial instruments (the target entity) is very important for understanding the investor's opinion.
For example, the closing price of a financial instrument is given every day, and it can provide a base for evaluating the degree of sentiment.
For instance, 303 and 307 in Figure~\ref{fig:Example of investor's opinion} cannot provide any information if we do not compare it with the closing price of \$AAPL.
When the closing price 297.32 is given, we can, therefore, infer the degree of sentiment as [1.91\%, 3.26\%] by a simple calculation.
The degree of sentiment acquired via this approach is more rational than those labels from $-$1 to 1 by the intuition of annotators in the previous work~\cite{cortis2017semeval}.
This kind of information not only provides the degree of sentiment but also implies the sentiment toward the target financial instrument.
Now the 7-tuple is extended to 8-tuple as follows: 
\begin{quote}
    $(e,s,h,t^p,a,d,C,M^e_{t^p})$\,,
\end{quote}
where $M^e_{t^p}$ denotes the market information set of target entity before publishing time.
The other research question is raised below:

\vspace{5pt}

\noindent (\textbf{RQ7}) How to align the information in the investor's claims to the market information of the target entity?

\vspace{5pt}

In most of opinion mining tasks, the opinions do not have the validity period.
However, since the financial market changes all the time, the investor's opinions do have a validity period, even the opinions of professional stock analysts are the same.
Most of the analysis reports of professional analysts set the validity period within one year even shorter.
Therefore, an investor's opinion can be represented as a 9-tuple:
\begin{quote}
     $(e,s,h,t^p,t^v,a,d,C,M^e_{t^p})$\,,
\end{quote}
where $t^v$ denotes the validity period.
Taking the instance in Figure~\ref{fig:Example of investor's opinion} as an example, the opinion is transferred into:
\begin{quote}
    (\$AAPL, Bullish, william, 1/3/20 11:44PM, 1/6/20--1/10/20 (this week), technical analysis, [1.91\%, 3.26\%], \{Price Target: [303,307]\}, \{Closing Price: 297.32\})
\end{quote}

The aforementioned lead to:
\vspace{5pt}

\noindent (\textbf{RQ8}) How long is the validity period of the investor's opinions?

\noindent (\textbf{RQ9}) Comparing with coarse-grained opinions, how much informativeness is increased for the downside tasks after capturing fine-grained opinions?

\section{Current Financial Opinion Mining Tasks}
\label{section opinion mining}
Although the opinion mining in the financial domain, the investor's opinions mainly, has been discussed for a long time, there are few benchmark datasets for reproducing the experimental results and making the extended researches.
In this section, we focus on sorting out the existing datasets of different tasks and the related state-of-the-art models, and further discuss the potential research questions.
Again, the datasets or task setting adopting market profit as labels such as StockNet~\cite{xu-cohen-2018-stock} will not be included, because these works do not capture any individual opinions.

\subsection{Tasks and Datasets}
Detecting the components in the opinion 9-tuple (7-tuple) defined in Section~\ref{section definition} is the challenge that has already been explored.
For the investor's sentiment, several works~\cite{li2017learning,chen2018ntusd}, adopt the labeled data from StockTwits directly, and do not publish the datasets for future research.
Therefore, these works are not comparable.
The dataset in Semeval-2017 task 5~\cite{cortis2017semeval} extends the investor's sentiment into the range from $-$1 to 1.
Total 2,499 financial tweets collected from Twitter and StockTwits are labeled by three annotators.
Jiang et al.~\shortcite{jiang-etal-2017-ecnu} perform the best with an ensemble model composed of support vector regression, XGBoost, AdaBoost, and bagging regressor.

For the aspect of the fine-grained investor's opinion, FiQA dataset~\cite{maia201818} provides 774 annotated tweets, and there are 4,847 annotated tweets in NumAttach dataset~\cite{chen2019numeral}.
E et al.~\shortcite{10.1145/3184558.3191825} perform the best in FiQA dataset with an attention-based LSTM model.
Since the annotation in NumAttach is used as the auxiliary task~\cite{chen2019numeral}, the performance on aspect detection is not yet explored.

Like the example in Figure~\ref{fig:Example of investor's opinion}, numerals are informative in the financial data.
However, the meanings of numerals are various.
In order to understand the fine-grained meaning of the numerals, we propose a taxonomy for the numerals in financial data, FinNum~\cite{chen2018numeral}.
There are five kinds of numerals related to investor's opinions, including price target, buying price, selling price, support price-level, and resistance price-level.
The informativeness and usefulness of these fine-grained opinions are already discussed in our previous work~\cite{chen2019crowd}.
The FinNum dataset provides 8,868 annotations on the numeral information on financial tweets.
This dataset can be used for detecting the numeral components in the investor's opinions.
Wang et al.~\shortcite{10.1007/978-3-030-36805-0_15} adopt BERT~\cite{devlin-etal-2019-bert} in this dataset, and gain the state-of-the-art performance.

The named entity recognition (NER) may not be a challenging task in financial data, because there are finite financial instruments.
As shown in Figure~\ref{fig:Example of investor's opinion}, the target entity (\$AAPL) is marked by the writer.
However, the linking between the opinion and the target entity is a challenge.
When analyzing the paragraph-level or document-level description, there may exist more than one financial instrument or financial product.
In the NumAttach dataset~\cite{chen2019numeral}, we show that even a tweet may mention more than one financial instrument.
There are 7,984 annotations for the linking between the target entity and the fine-grained opinion.

\subsection{Further Research Questions}
\label{Further Research Questions}
Since constructing a domain-specific dataset are costly, there are few benchmark datasets in financial opinion mining.
That also shows the chance for future research.
For example, the validity period detection and the fine-grained customer's opinion are not discussed yet.

Most of the previous works~\cite{bollen2011twitter,valencia2019price} adopt market movement prediction as the downside task of capturing investor's sentiment, and coarsely use the averaged sentiment score from all investors.
However, Wang et al.~\shortcite{10.1145/2675133.2675144} indicate that the top investors, ranking by their history performances, on social media platforms can achieve 75\% accuracy on market movement prediction.
For reference, the accuracy of the recent market movement prediction model~\cite{feng2019enhancing} is in the range of [53.05\%, 57.20\%].
Besides, as the opinions in other domains, with the financing incentive, some opinions may not be worthy of trust such as spam review.
In the financial domain, the exaggerate information~\cite{chen-etal-2019-numeracy} may influence the market, and the opinions with the exaggerate information may also be doubtful.
Therefore, detecting the doubtful opinions are also an open issue.
The above discussion arises the following research questions:
\vspace{5pt}

\noindent (\textbf{RQ10}) How to evaluate the quality of the investor's opinions?

\noindent (\textbf{RQ11}) What kinds of opinions are influential in finance?

\noindent (\textbf{RQ12}) What kinds of opinions are worth to truth in finance?

\section{Road Map of Future Research Issues}
\label{section road map}
\subsection{Argument Mining in Finance}
Argument mining is one of the focused topics in the AI community recently.
Cabrio and Villata~\shortcite{cabrio2018five} and Lawrence and Reed~\shortcite{lawrence2019argument} provide the surveys to the recent development of argument mining.
Please refer to their survey for details.
In our view, it can be considered as the next stage of fine-grained financial opinion mining.
Previous works and the above sections only focus on extracting the opinions of the investors or customers.
In this section, we discuss the importance of mining the premises and evaluating the rationales of a financial opinion.

In order to clarify the tasks, we use a passage (\textbf{E2}) selected from professional analysis as an example.
The target entity of (\textbf{E2}) is TSMC, and there are one fact (\textbf{F1}), three premises (\textbf{P1-3}), and one claim (\textbf{C1}) in (\textbf{E2}).

\vspace{5pt}

\noindent \textbf{Example (E2)}: 

\noindent \textit{\textbf{(F1)} The overall revenue of semiconductor industry 10--11/2018 is in line with expectations. 
As \textbf{(P1)} \uwave{the company's leading-edge in high-end process production continues to increase}, coupled with \textbf{(P2)} \uwave{Globalfoundries' withdrawal from competition} and \textbf{(P3)} \uwave{inconsistencies in Intel's process conversion}, we estimate that \textbf{(C1)} \textbf{TSMC's revenue in 4Q18 will approximate to 9.35 billion US dollars}.}

\vspace{5pt}

The first challenge of in-depth opinion analysis is that detecting the opinion and the rationales, i.e., the claim and the premise.
In the financial market, the investors debate on different financial instruments every day with different stances, bullish or bearish.
It just likes the situations where the debaters discuss different topics on the affirmative and negative sides.
In order to analyze the claims and the premises of the investors, the detection task is necessary.

Aiming to make the AI models becomes explainable, AI scientists strive for having proof and evidence for the models' predictions.
However, in the financial opinion mining field, people use all kinds of opinions directly without asking the reasons.
For example, should we give the same weight to the tweet ``\$TSMC Goooooo!'' and (\textbf{E2}) when analyzing the investors' opinions?
In most of the previous works, their weights are the same.
It shows that there is still room for future researches.

\begin{figure}[t]
    \centering
    \includegraphics[width=7cm]{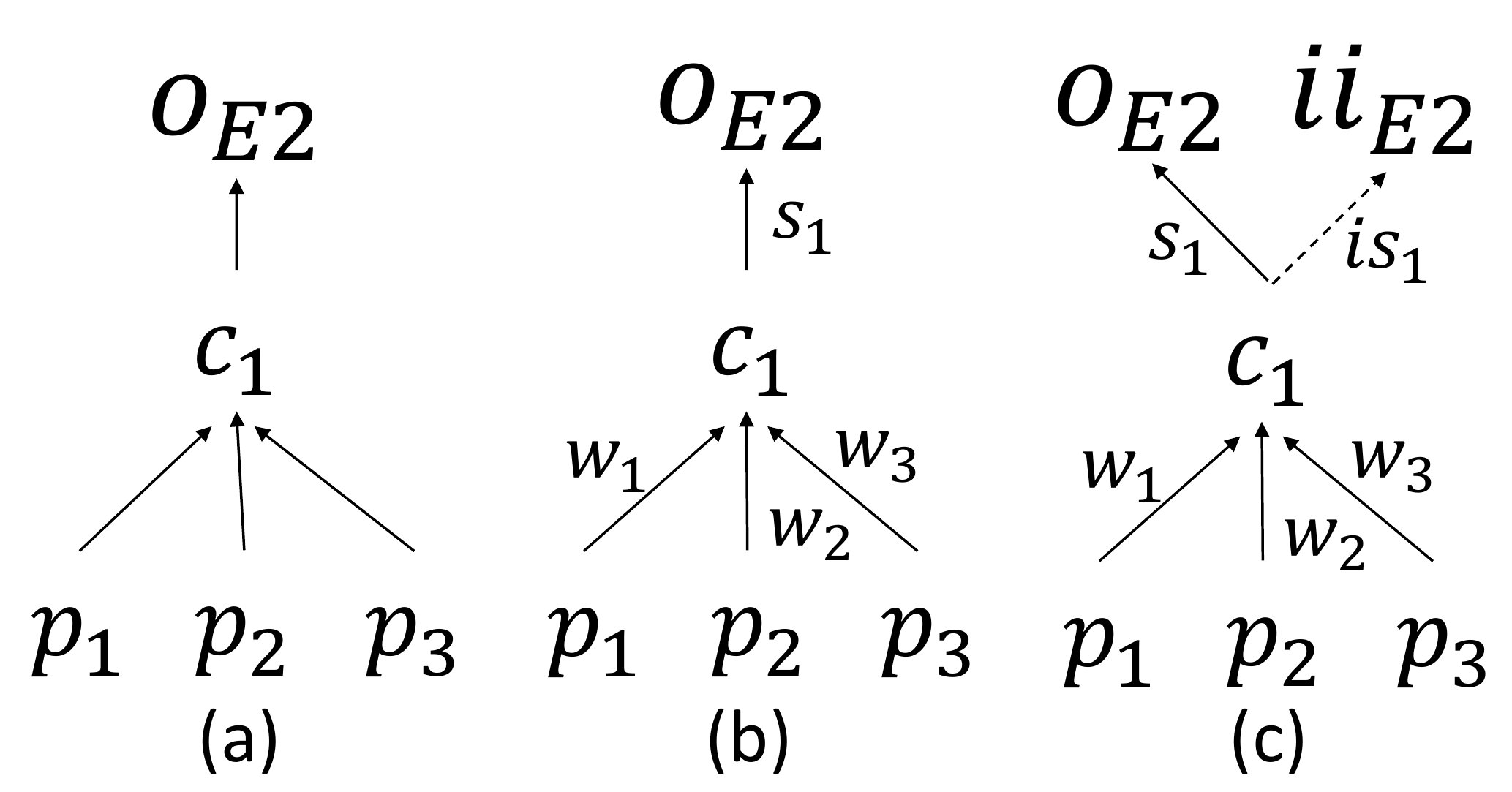}
    \caption{Directed graph of between financial opinion and arguments.}
    \label{fig:Micro opinion graph}
\end{figure}

One of the further research issues after claim and premise detection is relation linking.
In a narrative of an opinion, investors or customers may propose more than one claim with several premises.
After predicting the relationship, an opinion can be transformed into a graph as shown in Figure~\ref{fig:Micro opinion graph} (a).
Now, the claim set $C$ of an opinion may contain several premise, set $P$.

\subsection{Quality Evaluation}
After extracting the claims and the premises of an opinion, we can evaluate the opinion quality based on the extracted results.
Taking Figure~\ref{fig:Micro opinion graph} (b) for illustration, we can first evaluate the rationality of the premises, and we will get the rationality scores ($w_{1-3}$) of the premises.
We can further add up the rationality scores as the strength score ($s_1$) of the claim.
Then we can get the opinion quality ($q$) by accumulating the strength scores, and now an opinion can be represented as a 10-tuple or 8-tuple:
\begin{quote}
    $(e,s,h,t^p,t^v,a,d,C(P),M^e_{t^p},q)$
\end{quote}

\begin{quote}
    $(e,s,h,t^p,a,d,C(P),q)$
\end{quote}

In summary, this section provides a possible direction for (\textbf{RQ10}) in Section~\ref{Further Research Questions}.
That is, we can evaluate the opinions based on their claims and premises.

\subsection{Inferring Implicit Influence}
Different from other argument mining tasks, with the nature in the financial domain, we can infer the implicit influence (abbreviated as $ii$ hereafter) from an opinion.
That is, the bullish opinion of certain financial instruments could be bearish information of the other financial instrument, and vise versa.
This is also an issue of financial product/service.
We illustrate the implicit influence of (\textbf{E2}) in Figure~\ref{fig:Micro opinion graph} (c).
The claim (\textbf{C1}) in the example (\textbf{E2}) may also influence the other company in the semiconductor industry.
Therefore, we can infer the implicit influence ($ii_{E2}$) based on the influence score($is_1$).
The implicit influence can be represented as a 10-tuple as an opinion.

To sum, this section points out a research issue as follows:
\vspace{5pt}

\noindent (\textbf{RQ13}) How to infer the implicit influence embedded in an opinion?

\begin{figure}[t]
    \centering
    \includegraphics[width=8cm]{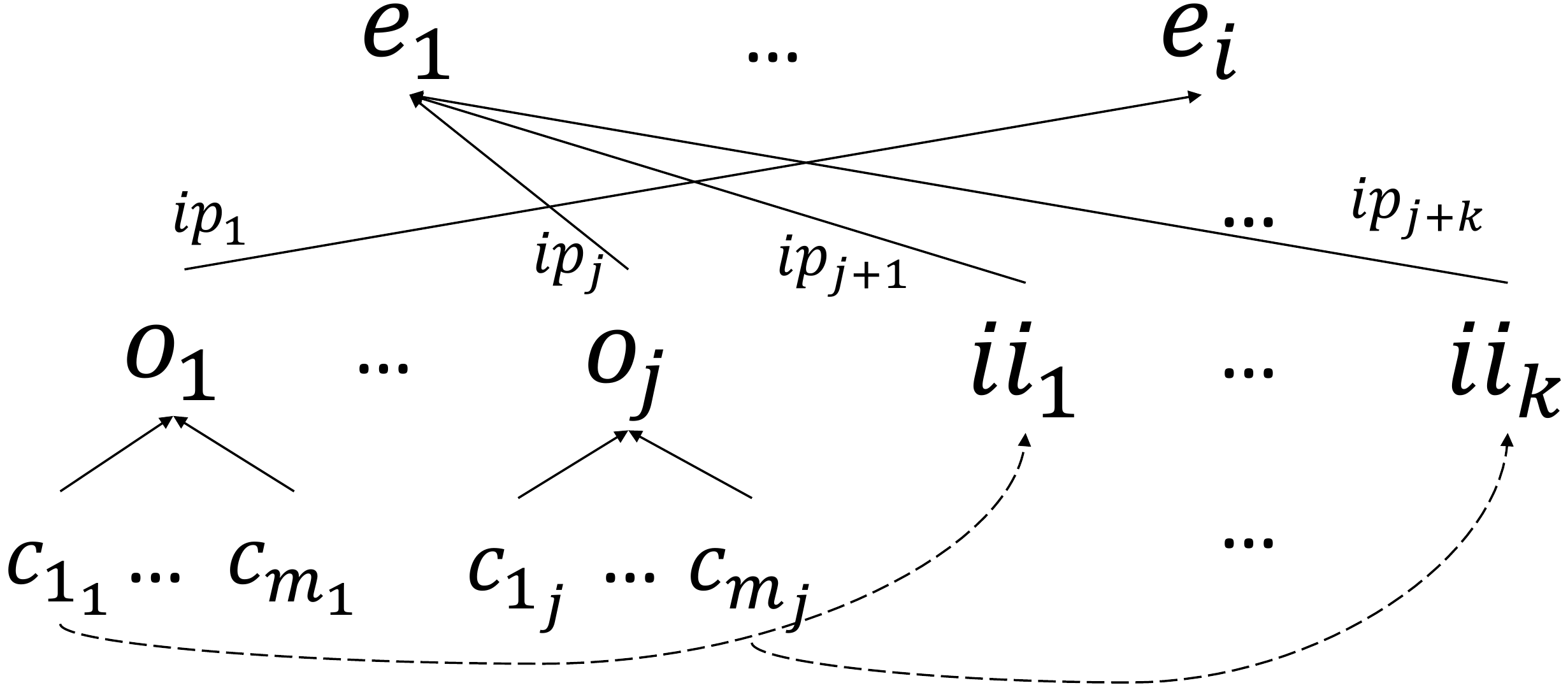}
    \caption{Directed graph between financial opinions and entities.}
    \label{fig:macro opinion graph}
\end{figure}

\subsection{Retrieval and Summarization}
Now, we complete the fine-grained opinion mining task on individual opinions.
The next stage is to compare the opinions and provide a global view of certain financial instruments/products/services.
We provide a directed graph in Figure~\ref{fig:macro opinion graph}.
Here, we use the case in financial instruments as an example.
Image that we are in the world with $i$ financial instruments ($e$), $j$ investors' opinions ($O$) with $k$ implicit influences ($ii$), and each investors' opinion has $m_j$ claims.
Each opinion node and implicit influence node have their own influence power ($ip$) toward certain financial instruments.
Now, an opinion can be shown as an 11-tuple or 9-tuple:

\begin{quote}
    $(e,s,h,t^p,t^v,a,d,C,M^e_{t^p},q,ip)$
\end{quote}

\begin{quote}
    $(e,s,h,t^p,a,d,C,q,ip)$
\end{quote}

Based on the thought, some research questions emerge:
\vspace{5pt}

\noindent (\textbf{RQ14}) How to evaluate the influence power of an opinion?

\noindent (\textbf{RQ15}) What is the relationship between the influence power and other components?

\vspace{5pt}

Leveraging to the opinion 11-tuple and 9-tuple, we can retrieve the opinions based on different kinds of queries.
For example, we can sort out the top 5 influential opinions of TSMC, which may be useful for analyzing the sales on June 2020 based on $(e,t^v,a,ip)$ in the opinion 11-tuple.

With the directed graph between the entities and the opinion 11-tuple (9-tuple), we can prune the graph based on different components. 
We can only adopt the opinions that may influence our analysis of the target entity, or those are important to downside tasks.

\begin{figure*}[t]
    \centering
    \includegraphics[width=18cm]{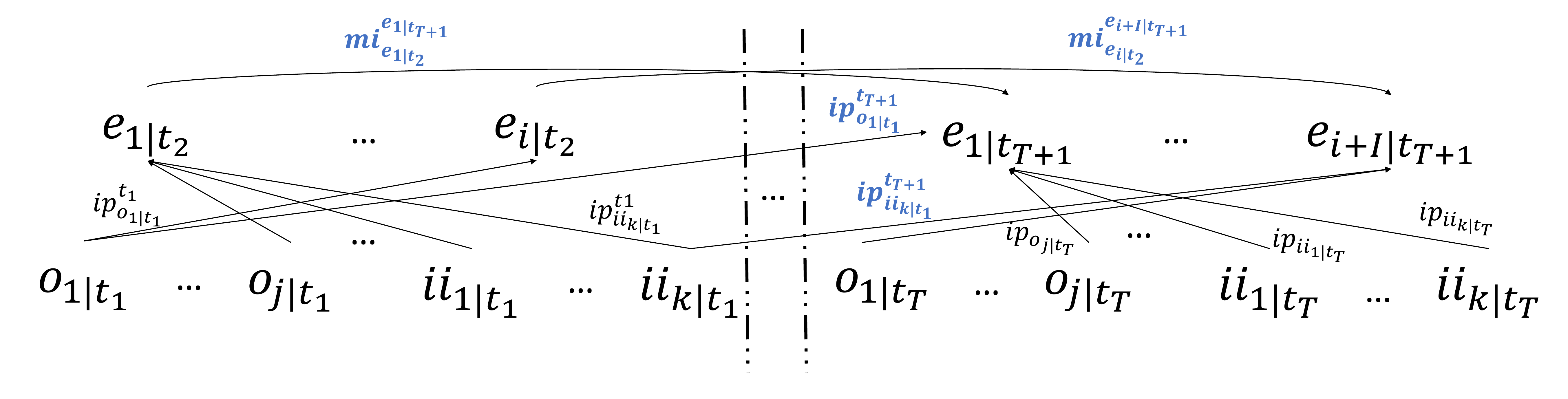}
    \caption{Directed graph in time series. The bold (blue) terms are the new terms that we discuss in Section~\ref{section tracing in time series}.}
    \label{fig:Graph of financial opinion in time series}
\end{figure*}

\subsection{Tracing in Time Series}
\label{section tracing in time series}
When analyzing financial data, time is one of the most important components that should be considered.
Till now, we only discuss the opinion at a certain time, i.e., $t^p$.
However, the opinion at time $t$ will not influence the status of the target entity at time $t$.
As shown in Figure~\ref{fig:Graph of financial opinion in time series}, the opinion at $t_1$ ($o_{1|t_1}$) may influence $e_1$ at time $T+1$ ($e_{1|t_{T+1}}$), and $ip^{t_{T+1}}_{o_{1|t_{1}}}$ is the influence power of $o_{1|t_1}$ to $e_{1|t_{T+1}}$.
Therefore, time is the condition of $ip$, and the $ip$ toward different entities at different time can be shown as a set $IP$.
Now, an opinion can be shown as follows:

\begin{quote}
    $(e,s,h,t^p,t^v,a,d,C(P),M^e_{t^p},q,IP)$
\end{quote}

\begin{quote}
    $(e,s,h,t^p,a,d,C(P),q,IP)$
\end{quote}

With the concept of time series, there are three kinds of opinions may exist in time $t+1$: (1) the new opinion in time $t+1$, (2) the opinion in time $t$ continuing to exist in time $t+1$, i.e, $t^v$ has not passed yet, and (3) the opinion in time $t$ changing at time $t+1$.
The opinion may be changed due to other opinions in time $t$.
That is, there exists an interaction between the opinions, and here arises the other research question:
\vspace{5pt}

\noindent (\textbf{RQ16}) How to evaluate or capture the interaction between the opinions?

\vspace{5pt}

The last interaction that we should consider is the one between $e_i$, i.e., the entities.
The status of $e_i$ at time $t$ may influence the status of itself at time $t+1$ and the status of other entities at time $t+1$.
The influence between entities is denoted as $mi$, market influence.
It can be linked to the research question in the microeconomics field.
Now, the overall picture from a financial opinion to the target entity is complete.

\section{Blueprint --- Solutions and Directions}
\label{section Blueprint}

\subsection{Component Extraction}
As we already showed, an opinion can be represented as an 11-tuple or a 9-tuple.
We can obtain some of the components in the 11-tuple (9-tuple) via information extraction techniques.
Because of many named entities, including $e$, $h$, $t^p$, and $t^v$, the NER task is the first step that we should consider.
For the target entity, as shown in Figure~\ref{fig:Example of investor's opinion}, investors (both professional analysts in the institution and individual investors) have idioms for the target entity such as the ticker of the financial instruments (AAPL US EQUITY) on the Bloomberg Terminal and the cashtag (\$AAPL) on StockTwits.
For financial products or services, the number of entities is finite.
Therefore, extracting the target entity may not be a challenge when dealing with financial textual data.

The identification of time expressions is more tackleable now.
However, since a time expression may have different meanings, disambiguating the validity period $t^v$ could remain a challenging task.
In the FinNum shared task~\cite{chen2019overview}, participants provide several useful features for understanding numeral information.
That may be useful for inspiring future researches.

\subsection{Relations of Components}
Some components could be inferred based on other components.
We list some examples as follows.
The sentiment ($s$) and the degree of sentiment ($d$) could be deduced from the comparison between market information ($M_{t^p}$) and the fine-grained claims ($c$) such as price target.
The opinion quality ($q$) could be analyzed based on the strength of the claims ($C$).
The influence power ($ip$) could be a function of the opinion holder ($h$) and the opinion quality ($q$).
For example, the tweet of Donald John Trump, the president of the United States, may have a higher $ip$ than that of a common person.
The implicit influence ($ii$) is also an interesting and complex research issue.
We may need to leverage the ontologies such as the Financial Industry Business Ontology (FIBO)\footnote{\url{https://spec.edmcouncil.org/fibo/index.html}} to construct the knowledge graph of the financial domain to deal with this issue.

\subsection{Entity Status Evaluation and Prediction}
Analyzing the status of $e_i$ is the final purpose of analyzing opinions.
The status of $e_i$ could be the credit cards in circulation, the sales of the insurance, the quality of the customer service, the price of the stock, or the market information of any financial instrument/product/service.
In Figure~\ref{fig:macro opinion graph}, we show that the current status of the entity can be formulated by the opinions related to it.
This information could be used for evaluating the current reputation of the entity.
As shown in Figure~\ref{fig:Graph of financial opinion in time series}, it could also be the cue for predicting the future status of the entity.

Based on the discussions in this paper, we suggest the researchers interested in this field pay more attention to completing the opinion 11-tuple (9-tuple) and the proposed graph of financial opinions, and further provide an in-depth analysis of the relations between the target entities and the components.
In this way, comparing with constructing an end-to-end model and focusing on the accuracy, things will become explainable and more rational.

\section{Conclusion}
\label{section conclusion}
This position paper provides a definition of fine-grained financial opinion and proposes the comprehensive directed graphs for real-world interaction between financial opinions and entities.
In addition to a survey of the existing datasets and tasks, this paper indicates 16 research questions for future works and provide feasible research directions for them.
We also indicate the important but untackled challenges in fine-grained financial opinion mining.
Our intent is to depict a big picture for researchers who involve in expediting the development of this topic.

\bibliographystyle{named}
\bibliography{ref}

\end{document}